\def\year{2019}\relax
\documentclass[letterpaper]{article} 
\usepackage{aaai19}  
\usepackage{times}  
\usepackage{helvet}  
\usepackage{courier}  
\usepackage{url}  
\usepackage{graphicx}  
\begin{document}

\title{NeuroX: A Toolkit for Analyzing Individual Neurons in Neural Networks}
\author{
  Fahim Dalvi,\textsuperscript{1}
  Avery Nortonsmith,\textsuperscript{2}
  Anthony Bau,\textsuperscript{2} \\
  {\bf \Large 
    Yonatan Belinkov,\textsuperscript{2}
    Hassan Sajjad,\textsuperscript{1}
    Nadir Durrani,\textsuperscript{1}
    James Glass\textsuperscript{2}
  } \\\\
  \textsuperscript{1}Qatar Computing Research Institute, HBKU Research Complex, Doha 5825, Qatar\\ 
  \textsuperscript{2}MIT Computer Science and Artificial Intelligence Laboratory, Cambridge, MA 02139, USA \\
  \texttt{\{faimaduddin,hsajjad,ndurrani\}@qf.org.qa} \\
  \texttt{\{averyn,abau,belinkov,glass\}@mit.edu} \\
}

\maketitle

\begin{abstract}

We present a toolkit 
to facilitate the interpretation and understanding of neural network models.  
The toolkit provides several 
methods to identify salient neurons with respect to the model itself or an external task. A user can visualize selected neurons, ablate them to 
measure their effect on the model accuracy, and manipulate them to control the behavior of the model at the test time. 
Such an analysis has a potential to serve as a springboard 
in various research directions, such as understanding 
the model, better architectural choices, model distillation and controlling data biases. The toolkit is available for download.\footnote{\url{https://github.com/fdalvi/NeuroX}}
\end{abstract}

\section{Introduction}
%
Despite the remarkable evolution of deep neural networks in natural language processing, their interpretability remains a challenge. 
In this work, we aim to facilitate analysis of neural models by introducing the \texttt{NeuroX} toolkit. 
The toolkit enables users to analyze models, 
and is flexible to use with any 
neural model. Given a trained model, the toolkit provides several mechanisms to probe it at neuron level. 
A user can visualize individual/group of neurons, ablate them to see the effect on the overall quality, and manipulate them to control the behavior of the model at test time.
The analysis provided by the toolkit
opens new horizons of research in various directions, such as understanding the internal learning of the model, better architectural choices, model distillation,  controlling data biases, etc.


\begin{figure*}[ht]
\centering
\includegraphics[width=0.90\linewidth]{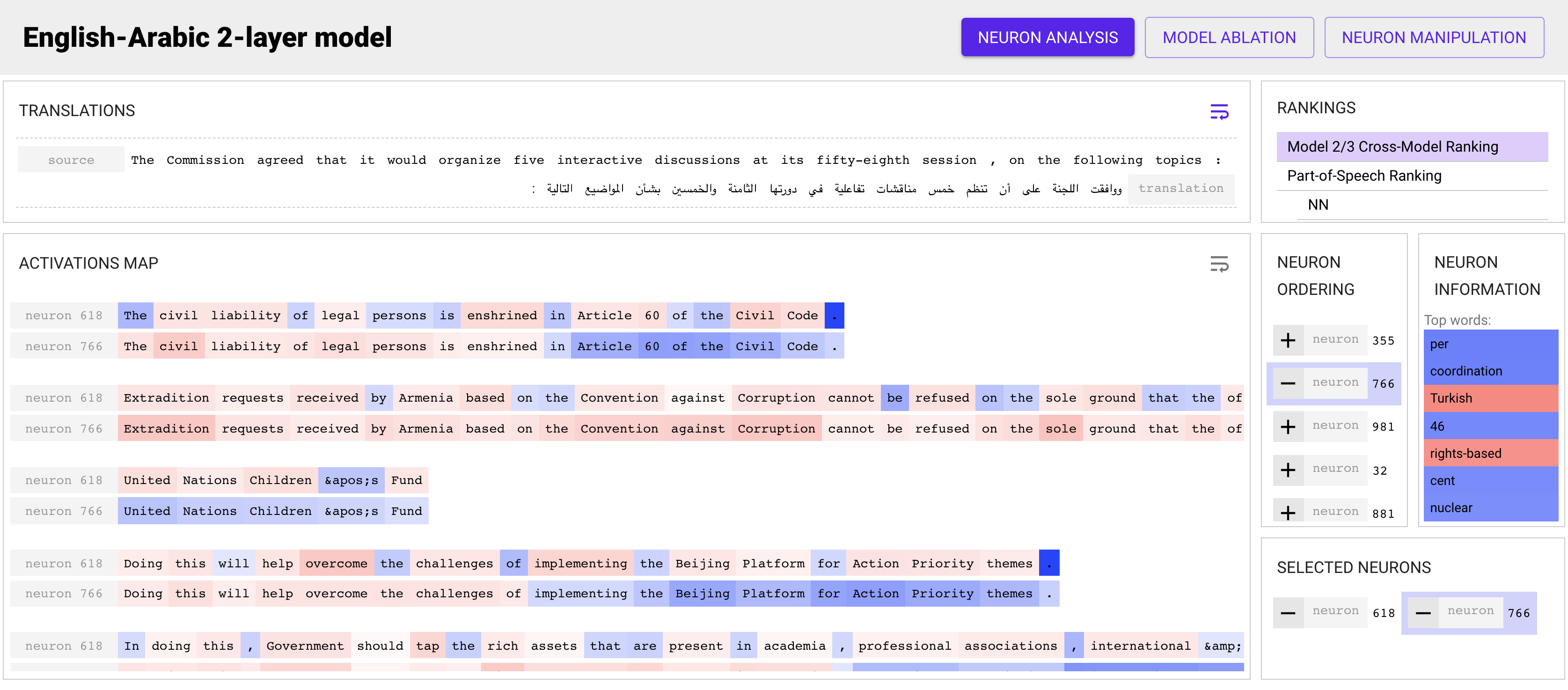}
\caption{Analysis Screen}
\label{fig:analysisScreen}
\end{figure*}

\section{Analysis Methods}
Given a trained model (or a set of models), the toolkit extracts the activations 
from the model(s) and runs several analyses routines on it. 
It then provides an interactive visual interface to analyze 
neurons.
We provide three 
methods to analyze neural network models in the toolkit: \texttt{Individual and Cross-model Analysis}, 
to 
search for neurons, important for the model itself \cite{indivdualneuron:iclr} and 
\texttt{Linguistic Correlation Analysis}, which identifies important neurons w.r.t. an extrinsic property \cite{dalvi_research:aaai19}. 
The output of each method is a ranked list of neurons. 

\paragraph{Individual Model Analysis} provides an intrinsic way to comprehend the quality of a model. It enables the user to visualize the effect of various design choices, such as layer normalization and dropout, on individual neurons, and can lead towards improving the architectural design of the model. For example, given two similar models, one trained with and another without layer normalization, one can visualize the effect of layer normalization on model activations using various provided ranking strategies, such as variance, distance from mean, etc.

\paragraph{Cross-model Analysis} provides a ranking of neurons based on the correlation among several models. Such rankings provides an insight of the prevalent properties that the models focused on while learning the task. For instance, in the task of machine translation, 
we found that neurons handling the length of the sentence and end of sentence are the most salient neurons across the models \cite{dalvi_research:aaai19}. 

\paragraph{Linguistic Correlation Analysis} enables a user to analyze neurons 
against any external task.
A task can be as simple as predicting \emph{position of a word} in a sentence, or {months of year} tag 
or as complex as 
predicting the morphological or semantic property of a word. The task-specific analysis helps to connect the dots between the features learned automatically by a neural model and features manually extracted for a non-neural network model. It also enables to answer questions like
what has a neural model learned about word morphology and lexical semantics and finding neurons that are explicitly focusing at a specific phenomenon.

\section{Visualization}
Given the identified salient neurons, a user can select individual or groups of neurons and visualize them in action on the input text. In addition, a user sees statistics about the selected neurons and a list of top words where the selected neurons are most activated. 
Visualization is especially helpful in analyzing activations of the model independent of any analysis methodologies. It spurs 
researchers to look for questions whose answers are not straightforward to define as a learning task. 

\section{Neuron Ablation}
Masking-out of neurons has become a standard practice to probe neural network models~\cite{s.2018on}. With a number of ranking strategies provided, a user can 
quickly find the importance of different neurons in the model by ablating them. 
This 
could facilitate model distillation where 
the size of the model can be reduced by removing the least salient neurons with minimum deterioration in quality.

\section{Neuron Manipulation}

A user can identify neurons specific to a phenomenon either manually (via visualization) or automatically (via task-specific analysis) using the toolkit. Neuron manipulation \cite{indivdualneuron:iclr} enables the user to 
control
the output of the model at test time. This is useful for the purpose of improving the model and to reduce the bias learned from the data. For example, gender bias is a well known issue in most available datasets. With the help of the task-specific analysis, a user can identify neurons responsible for gender and manipulate their values at test time to output the desired gender.

\section{Use Cases}
Figure \ref{fig:analysisScreen} shows an example of the analysis provided by the toolkit. The screen is divided into several views. The top left part shows the source and translated text. The bottom left part shows visualization of selected neurons (an end-of-sentence marker neuron and a position neuron).
On the top right, one can see the list of analyses selected by the user to process. Based on the selected option, the interface shows a list of ranked neurons. Upon selecting a neuron (or multiple neurons), the user sees the list of top words the neuron is activated on and also visualizes it on the test set. 
In addition, the user can ablate and manipulate the selected neurons and see the effect on the translated output and on the overall score of the model.

\section{Related Work}
Google's \texttt{What-If} 
tool 
inspects machine learning models and provides users an insight of the trained model based on the predictions. In comparison, our toolkit is focused on deep learning models, and it provides several 
methods to analyze activations of a neural model.
\texttt{Seq2Seq-Vis} toolkit~\cite{seq2seq-vis} 
is a tool to debug NMT systems that enables the user to trace back the prediction decisions to the input of the model. Different from them, 
we focus on the internal understanding of the model, i.e., individual neurons, and on tracing linguistic information in the neurons. 
%
The toolkit can be used to analyze any neural network model and is flexible to integrate new analysis strategies.



\bibliography{naaclhlt2018,acl2017,thesis}
\bibliographystyle{aaai}
\end{document}